\documentclass[conference]{IEEEtran}
\usepackage{graphicx}
\usepackage{epstopdf}
\usepackage{cite}
\usepackage{amsmath,amssymb,amsfonts}
\usepackage{algorithmic}
\usepackage{textcomp}
\usepackage{tikz}
\usepackage{xcolor}
\usepackage{amsmath}
\usepackage{hyperref}
\usepackage{float}
\usepackage{caption} 
\newcommand{\squeezeup}{\vspace{-2.5mm}}

\def\BibTeX{{\rm B\kern-.05em{\sc i\kern-.025em b}\kern-.08em
    T\kern-.1667em\lower.7ex\hbox{E}\kern-.125emX}}
\begin{document}

\title{Reinforcement Learning Controlled Adaptive PSO for Task Offloading in IIoT Edge Computing \\
}

\author{\IEEEauthorblockN{Minod Perera, Sheik Mohammad Mostakim Fattah, Sajib Mistry, Aneesh Krishna} 
\IEEEauthorblockA{\textit{Faculty of Science and Engineering — School of EECMS, Curtin University, Perth, Australia} \\
Emails: m.perera11@student.curtin.edu.au, Sheik.Fattah@curtin.edu.au, Sajib.Mistry@curtin.edu.au,\\Aneesh.Krishna@curtin.edu.au}
}

\maketitle

\begin{abstract}
Industrial Internet of Things (IIoT) applications demand efficient task offloading to handle heavy data loads with minimal latency. Mobile Edge Computing (MEC) brings computation closer to devices to reduce latency and server load, optimal performance requires advanced optimization techniques. We propose a novel solution combining Adaptive Particle Swarm Optimization (APSO) with Reinforcement Learning, specifically Soft Actor Critic (SAC), to enhance task offloading decisions in MEC environments. This hybrid approach leverages swarm intelligence and predictive models to adapt to dynamic variables such as human interactions and environmental changes. Our method improves resource management and service quality, achieving optimal task offloading and resource distribution in IIoT edge computing.
\end{abstract}


\section{Introduction}
The advent of 5G networks has intensified the transmission of large data volumes, including high-definition videos, augmented reality (AR), and virtual reality (VR), all demand low latency, high bandwidth, and reliable connections. Processing this data at the network's central core introduces latency issues for time-sensitive tasks. Mobile Edge Computing (MEC) \cite{8030322} mitigates this by offloading compute-intensive tasks to proximate servers, reducing latency and server load. However, mobile devices often lack the processing power to fully utilize this capability.

\begin{figure}
    \centering
    \includegraphics[width=0.65\linewidth]{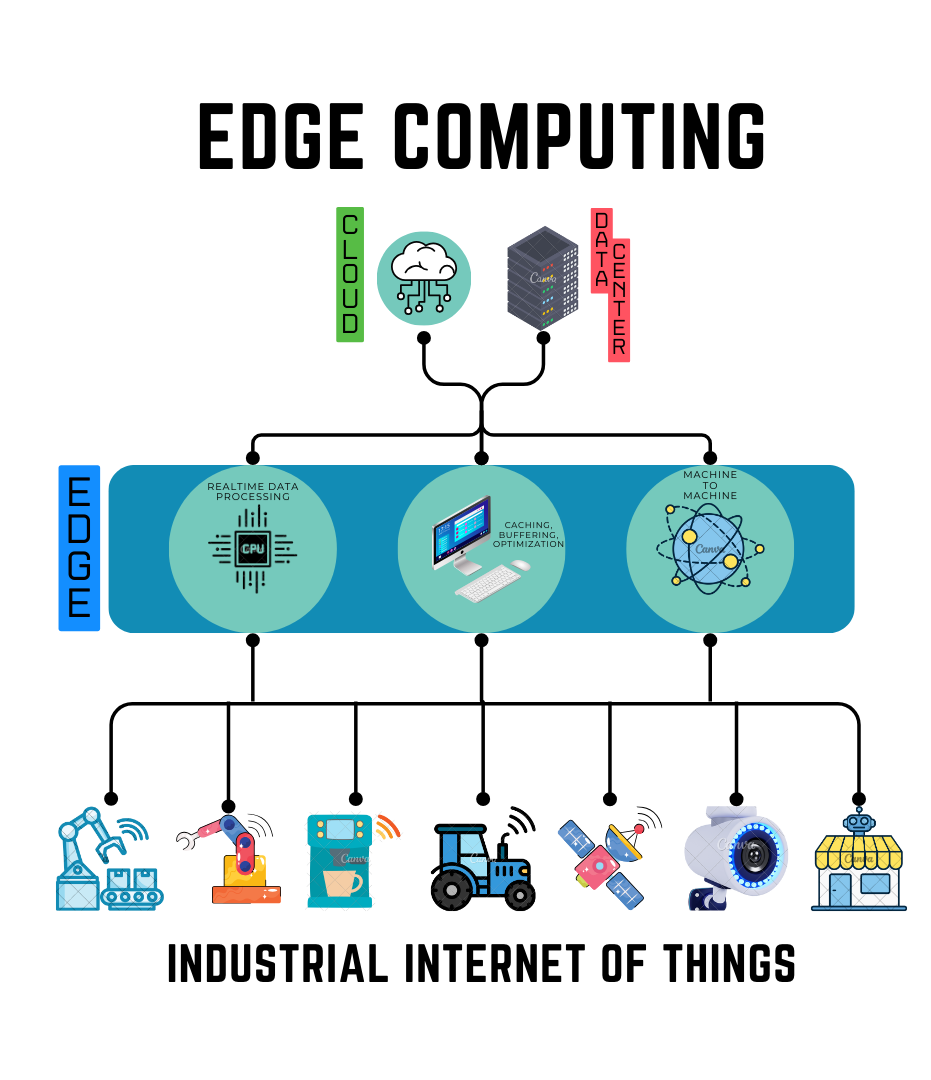}
    \caption{Environment of Smart Factory}
    \label{EnvSmart}
\end{figure}

In Industrial Internet of Things (IIoT) environments such as smart factories [Fig. \ref{EnvSmart}], efficient task offloading is crucial to enhance the production processes through data collection and analysis. Selecting the optimal MEC server for task offloading is a complex problem influenced by factors such as server processing capabilities, current load, and environmental changes. Traditional methods like always choosing the nearest server are suboptimal.

Particle Swarm Optimization (PSO) has been shown to be effective for task offloading in relatively static environments but struggles as the number of servers and devices increases. Reinforcement Learning (RL) algorithms, particularly in dynamic environments, offer promising solutions. This paper proposes a novel approach that integrates Soft Actor Critic (SAC), a type of RL algorithm, with Adaptive Particle Swarm Optimization (APSO) to enhance task offloading efficiency in MEC for IIoT environments.

\section{Related Work}

Mobile Edge Computing (MEC) addresses the limitations of centralized cloud computing by distributing computational workloads closer to data sources, reducing latency and server strain in IoT environments. The role of MEC in enhancing cloud and IoT systems is discussed in \cite{AAZAM2018278}. Energy efficiency in task offloading is improved using differential evolution, as shown in \cite{Sun2021}.

Particle Swarm Optimization (PSO) has been applied effectively to task offloading due to its simplicity and efficiency. Energy-efficient service allocation in fog computing using PSO and other meta-heuristics is presented in \cite{Mishra2018}. PSO is demonstrated to outperform genetic algorithms and simulated annealing in reducing energy consumption and latency in multi-user, multi-server IIoT environments in \cite{you2021efficient}. Further optimizations of PSO for IIoT task offloading achieve reduced latency and energy consumption, as described in \cite{Li2022}.

Reinforcement Learning (RL) has also been employed to manage dynamic resource allocation in MEC. The application of Deep Deterministic Policy Gradient (DDPG) to minimize task delay in IoT systems is detailed in \cite{chen2020}. An RL-based approach for UAV-assisted task offloading effectively handles high-dimensional state and action spaces, as proposed in \cite{Xu2022}. However, these RL methods may not be optimal for IIoT environments with mostly static device and server placements.

This work combines the strengths of PSO and RL to improve task offloading in MEC for IIoT environments. By integrating Soft Actor Critic (SAC) with Adaptive PSO, the limitations of both methods are addressed, enhancing efficiency in scenarios with known search spaces and static placements.

\section{Methodology}

We propose a novel approach that integrates Adaptive Particle Swarm Optimization (APSO) with Soft Actor Critic (SAC) reinforcement learning to enhance task offloading decisions in MEC environments.

Particle Swarm Optimization (PSO) is a population-based optimization algorithm inspired by the social behavior of flocks of birds or schools of fish. Particles represent potential solutions and adjust their positions based on personal and global best experiences.  Each generated particle, $i$, at any time-step $t$, represents a solution to the given problem and has four characteristics, its position $(x_i(t))$, velocity $(v_i(t))$, personal-best solution $(pbest)$, and current fitness $(f(x^t_i))$.

\begin{equation} \label{PSOeq}
    \begin{aligned}    
        v_i(t+1) = w_t*v_i(t)+c^1_t*U(a,b)(pbest_i(t) - x_i (t)) \\
        + c^2_t*U(a,b)(gbest (t) - x_i (t))
    \end{aligned}
\end{equation}

Equation \ref{PSOeq} showcases how PSO performs a simulation step. Each particle's position, i.e. the solution, is updated using their respective velocity vector. This process is repeated until a termination criterion is met, in general a number of iterations is chosen, and in our case 50 iterations will be the threshold be termination occurs.  PSO is effective in static environment's, but its performance degrades in dynamic scenarios due to fixed hyperparameters ($c^1_t, c^2_t, \And w(t)$).

To overcome this limitation, Adaptive-PSO (APSO) dynamically adjusts the PSO hyperparameters during the optimization process to enhance exploration and exploitation capabilities. The concept of Adaptive Particle Swarm Optimization (APSO) was introduced as an extension to PSO, enhancing the algorithm with a mechanism to dynamically change its hyperparameters during the optimization process \cite{ZhanAPSO}. The Evolutionary Factor, which reflects the swarm’s distribution state, enables transitions between Convergence, Exploitation, Exploration, and Jumping Out states. This dynamic adjustment allows APSO to gradually shift between these states, optimizing hyperparameters more effectively to achieve the optimal result.

However, APSO relies on problem-specific adaptation strategies, requiring extensive domain knowledge and manual tuning. We enhance base PSO to mimic APSO, by integrating Soft Actor Critic (SAC), an off-policy reinforcement learning algorithm that optimizes a stochastic policy by maximizing expected reward while encouraging exploration through entropy regularization. SAC balances exploration and exploitation by adjusting the policy based on both the reward and the entropy of actions, and was chosen for it's insensitivity to hyper-parameter tuning, allowing the algorithm to be used without fine tuning and extensive domain knowledge requirements.

\begin{equation} \label{SACobjfunc}
    \begin{aligned}
        \pi^* = \arg \max_{\pi} \mathbb{E}_{\tau \sim \pi} \Bigg[ \sum_{t=0}^{\infty} \gamma^t \Big( R(s_t, a_t, s_{t+1}) \\
        + \alpha H(\pi(\cdot | s_t)) \Big) \Bigg]
    \end{aligned}   
\end{equation}

where:
\begin{itemize}
    \item \( R(s_t, a_t, s_{t+1}) \) is the reward for taking action \( a_t \) in state \( s_t \),
    \item \( \gamma \) is the discount factor,
    \item \( H(\pi(\cdot | s_t)) = \mathbb{E}_{a \sim \pi(\cdot | s_t)} [-\log \pi(a | s_t)] \) is the entropy of the policy at state \( s_t \),
    \item \( \alpha \) is the entropy coefficient, which controls the trade-off between exploration and exploitation.
\end{itemize}

Equation \ref{SACobjfunc} shows the objective function in SAC, which combines the expected reward with an entropy bonus, promoting exploration by favoring more diverse actions. 

In our approach, SAC is used to adaptively fine-tune the hyperparameters of PSO in response to environmental changes. The SAC agent observes the performance of PSO and adjusts the acceleration coefficient parameters to improve convergence and adaptability. This integration allows the system to handle dynamic variables and scaling in IIoT environments without manual intervention. The State-Value Function $V^\pi(s)$, which is a measure of the expected return or the cumulative reward the agent can achieve in a given state $s$, includes the entropy bonus at each timestep.

\begin{equation} \label{Sfunc}
    \begin{aligned}
        V^{\pi}(s) = \mathbb{E}_{\tau \sim \pi} \Bigg[ \sum_{t=0}^{\infty} \gamma^t \Bigg( R(s_t, a_t, s_{t+1}) \\
        + \alpha H(\pi(\cdot | s_t)) \Bigg) \Bigg| s_0 = s \Bigg]
    \end{aligned}
\end{equation}

The State-Action Value Function allows the agent to make decision on choosing what actions to follow in a given state $s$, by measuring the expected return from taking a specific action \( a \), while continuing to follow the policy \( \pi \) afterward.

By combining SAC with APSO, our method leverages the strengths of both algorithms: PSO's efficient exploration of the solution space and SAC's ability to learn optimal strategies in dynamic settings. This hybrid approach improves task offloading efficiency, reduces latency and energy consumption, and adapts to changes in the MEC environment.

\section{Experimental Results}

To evaluate the effectiveness of our proposed APSO-SAC method, we conducted experiments measuring the best cost achieved in task-offloading scenarios. The best cost combines both the computational cost and the latency incurred during the offloading process, providing a comprehensive measure of the efficiency of the algorithm.

In our simulation environment created with Python and Matlab, we consider 250 devices and 20 MEC servers, obtaining the best cost from a 10-run average. Each device has attributes such as data size, computational workload (completion requirement), and RAM requirement, while each server is characterized by processing speed, cost per unit time, and available RAM. The parameters are randomly generated within realistic ranges to simulate a diverse set of devices and servers.

The implementation details, code, and datasets for this work are available on \href{https://github.com/MinodRashmika/APSO_SAC_IIOT}{GitHub}

\subsection{Experiment Settings}

The parameters were to randomly generate the servers, devices, as well as tasks, in given ranges, as well as the hyperparameters of the RL algorithms, to create the simulated environment.

\subsubsection{Hyperparameters}

The setting used with the RL algorithms to fine-tune them for the given MEC in IIoT environment can be seen as follows. Further tuning of these hyperparameters can possibly yield greater results but will require further domain knowledge as well as extensive testing.

\begin{table}[H]
\centering
\captionsetup{skip=10pt}
\begin{tabular}{|l|c|}
    \hline
    \textbf{Hyperparameter} & \textbf{Value} \\
    \hline
    Optimizer & Adam \\
    Actor Learning Rate & $3 \times 10^{-4}$ \\
    Critic Learning Rate & $3 \times 10^{-4}$ \\
    Replay Size & 1,000,000 \\
    Update Rate ($\tau$) & 5e-3 \\
    Discount Factor ($\gamma$) & 0.99 \\
    Batch Size & 256 \\
    Loss Function & Mean Squared Error (MSE) \\
    Action Noise & Gaussian \\
    Training Steps & 1,000,000 \\
    \hline
\end{tabular}
\hfill\\
\caption{Hyperparameters for SAC}
\label{tab:sac_hyperparameters}
\end{table}

\subsubsection{Datasets}

The data ranges that are randomly set, decided by a given seed for consistency among testing suites can be seen as follows.  

\begin{table}[H]
\centering
\captionsetup{skip=10pt} 
\begin{tabular}{|l|c|}
    \hline
    \textbf{Parameter} & \textbf{Range} \\
    \hline
    Network Speed & 60-900 Mbps \\
    Server Cost & \$0.02-0.06/second \\
    Server Speed & 10-200 MI/second \\
    RAM & 2-8 GB \\
    Data Size & 50-150 MB \\
    Completion Requirement & 20-40 MI \\
    RAM Requirement & 1-2 GB \\
    $m$ (Cost Weight) & 10 \\
    $n$ (Latency Weight) & $1 \times 10^{-2}$ \\
    \hline
\end{tabular}
\caption{Parameter Ranges for MEC Offloading Optimization}
\label{tab:mec_parameters}
\end{table}

\subsubsection{Evaluation Techniques}

The total cost for a given task offloading configuration (i.e., the assignment of devices to servers) is calculated using the following equation:

\[
\text{Total Cost} = \sum_{j=1}^{N_{\text{devices}}} \left[ m \cdot C_{j} + n \cdot T_{j} \right]
\]

where:

- \( N_{\text{devices}} \) is the total number of devices.
- \( m \) and \( n \) are weighting factors for the cost and latency components, respectively.
- \( C_{j} \) is the computational cost for device \( j \), calculated as:
  \small
  \[
  C_{j} = \begin{cases}
  \text{ServerCost}_{s_j} \times T_{j}, & \text{if } \text{RAMRequirement}_{j} \leq \text{ServerRAM}_{s_j} \\
  0, & \text{otherwise}
  \end{cases}
  \]
  \normalsize 
- \( T_{j} \) is the total time for device \( j \), including both data transfer time and processing time:
  \[
  T_{j} = \frac{\text{DataSize}_{j}}{\text{NetworkSpeed}_{j}} + \frac{\text{CompletionRequirement}_{j}}{\text{ServerSpeed}_{s_j}}
  \]
- \( s_j \) is the server assigned to device \( j \).

In this calculation:

\begin{itemize}
    \item\textbf{Data Transfer Time} (\( \frac{\text{DataSize}_{j}}{\text{NetworkSpeed}_{j}} \)) represents the time required to transfer data from the device to the server.
    \item\textbf{Processing Time} (\( \frac{\text{CompletionRequirement}_{j}}{\text{ServerSpeed}_{s_j}} \)) represents the time the server takes to process the task.
    \item\textbf{Computational Cost} (\( \text{ServerCost}_{s_j} \times T_{j} \)) is the cost incurred by the server to perform the task over time \( T_{j} \).
    \item The conditional in \( C_{j} \) ensures that a device is only assigned to a server if the server has sufficient RAM to handle the device's RAM requirement.
\end{itemize}

We set the weighting factors \( m = 10 \) and \( n = 0.01 \) to balance the impact of cost and latency on total cost.

We perform multiple runs of the PSO algorithm, recording the best cost achieved in each run. The \textbf{best cost} represents the minimum total cost found by the optimizer for a particular server assignment strategy. By comparing the best costs across different methods, standard PSO, adaptive PSO, and our proposed APSO-SAC, we assess the effectiveness of our approach in minimizing task offloading costs in MEC environments.

\subsection{Implementations of Computational Offloading}

\subsubsection{Particle Swarm Optimization (Baseline)}

\begin{figure}[H]
    \centering
    \includegraphics[width=2.8in]{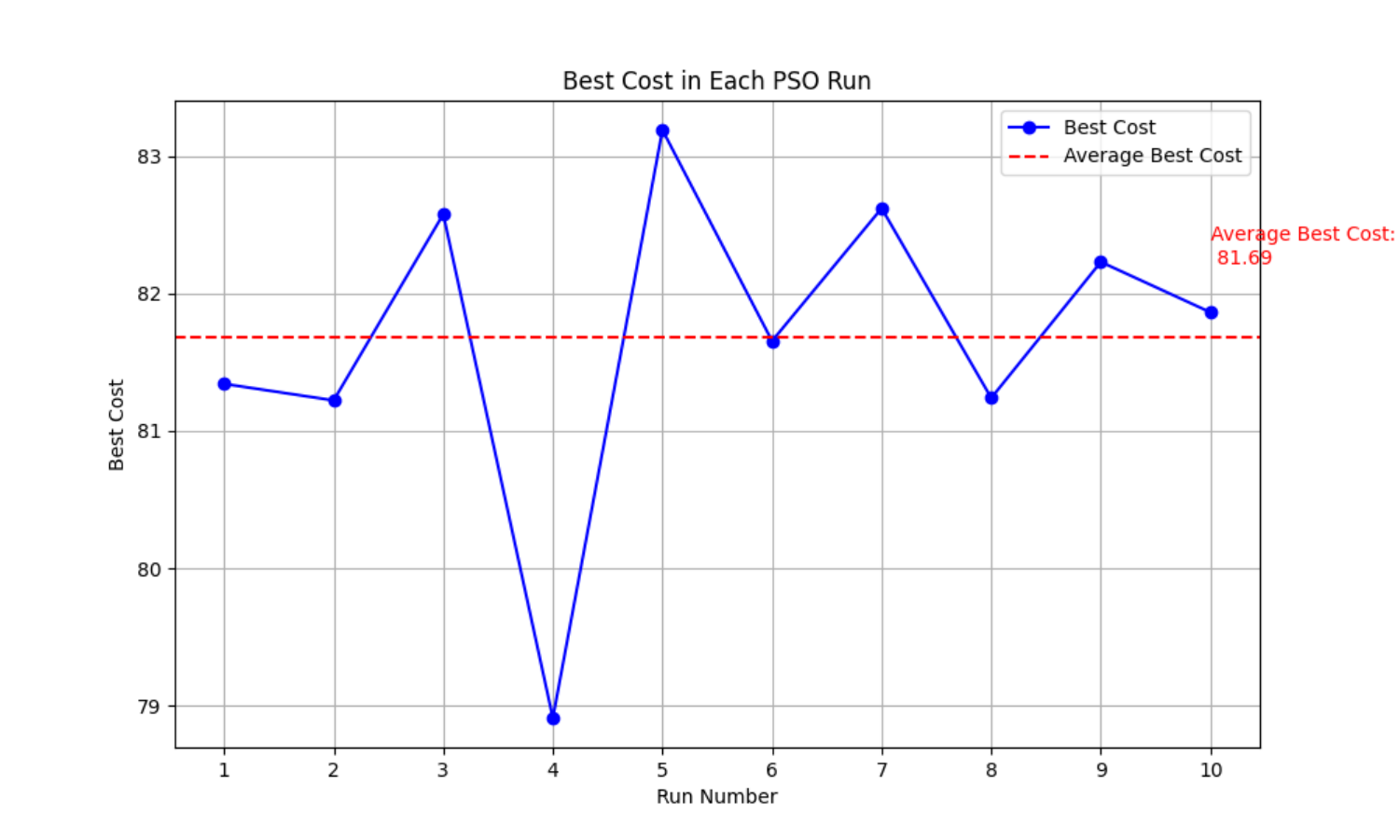}
    \caption{Baseline PSO}
    \label{psobaseline}
\end{figure}

Figure \ref{psobaseline} shows the baseline result from running an instance of PSO in our simulated large-scale IIoT environment, showing an average best cost of 81.69.
\squeezeup
\squeezeup
\subsubsection{Adaptive Particle Swarm Optimization}

Figure \ref{apso} shows the result of running an instance of Adaptive PSO in our simulated environment, showing an average cost of 76.39, while this is an improvement over base PSO, there is an exponential increase in runtime, up to a staggering 110\%, nullifying any benefits gained from the 6. 5\% reduction in the best cost.
\squeezeup
\begin{figure}[H]
    \centering
    \includegraphics[width=2.8in]{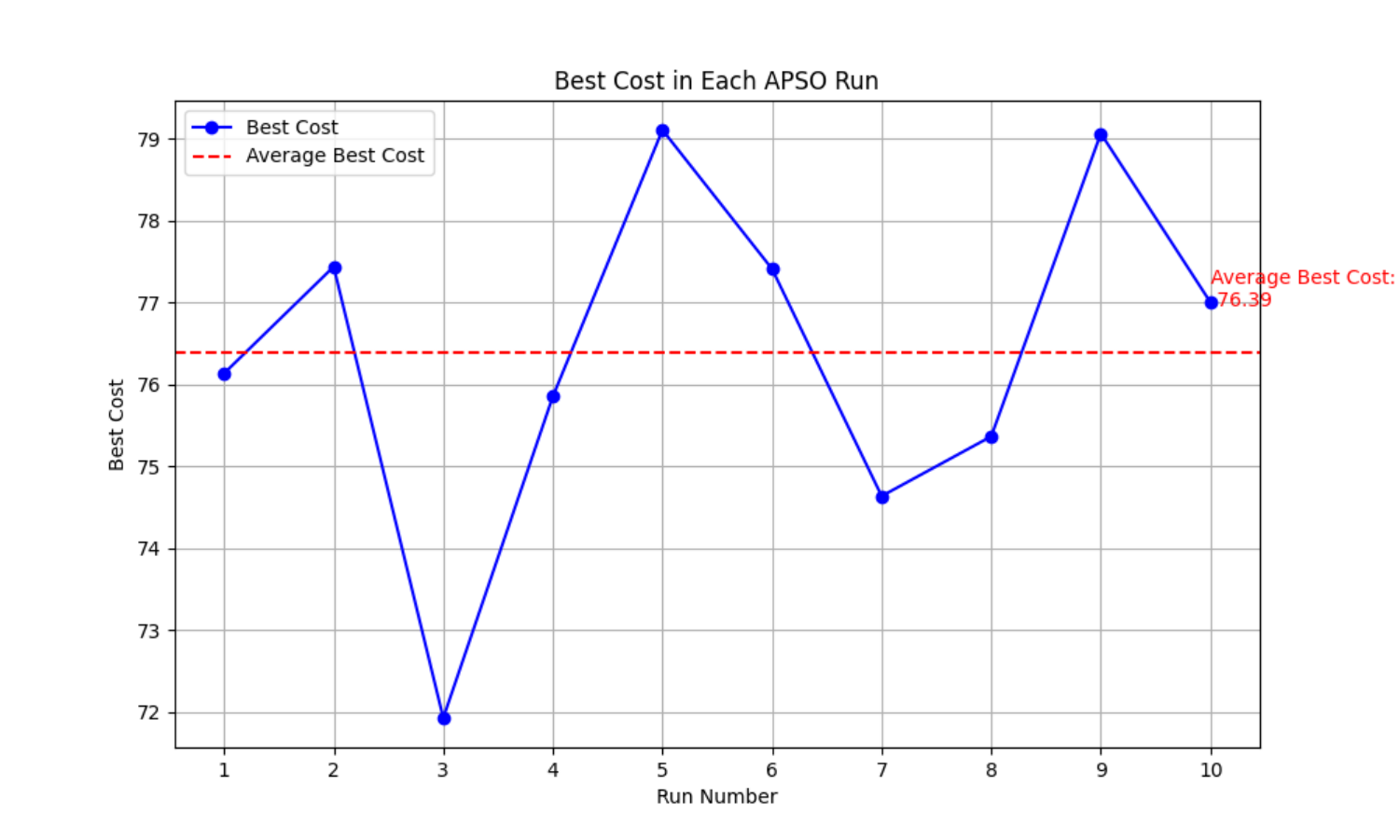}
    \caption{Adaptive PSO}
    \label{apso}
\end{figure}
\squeezeup
\subsubsection{Novel Approach - APSOSAC}

Figure \ref{apso} shows the result of running an instance of Adaptive PSO in our simulated environment, showing an average cost of 58.5, showing an improvement of 28.38\% over base PSO, while maintaining the same runtime, with the only caveat being the initial training period.
\squeezeup
\begin{figure}[H]
    \centering
    \includegraphics[width=2.8in]{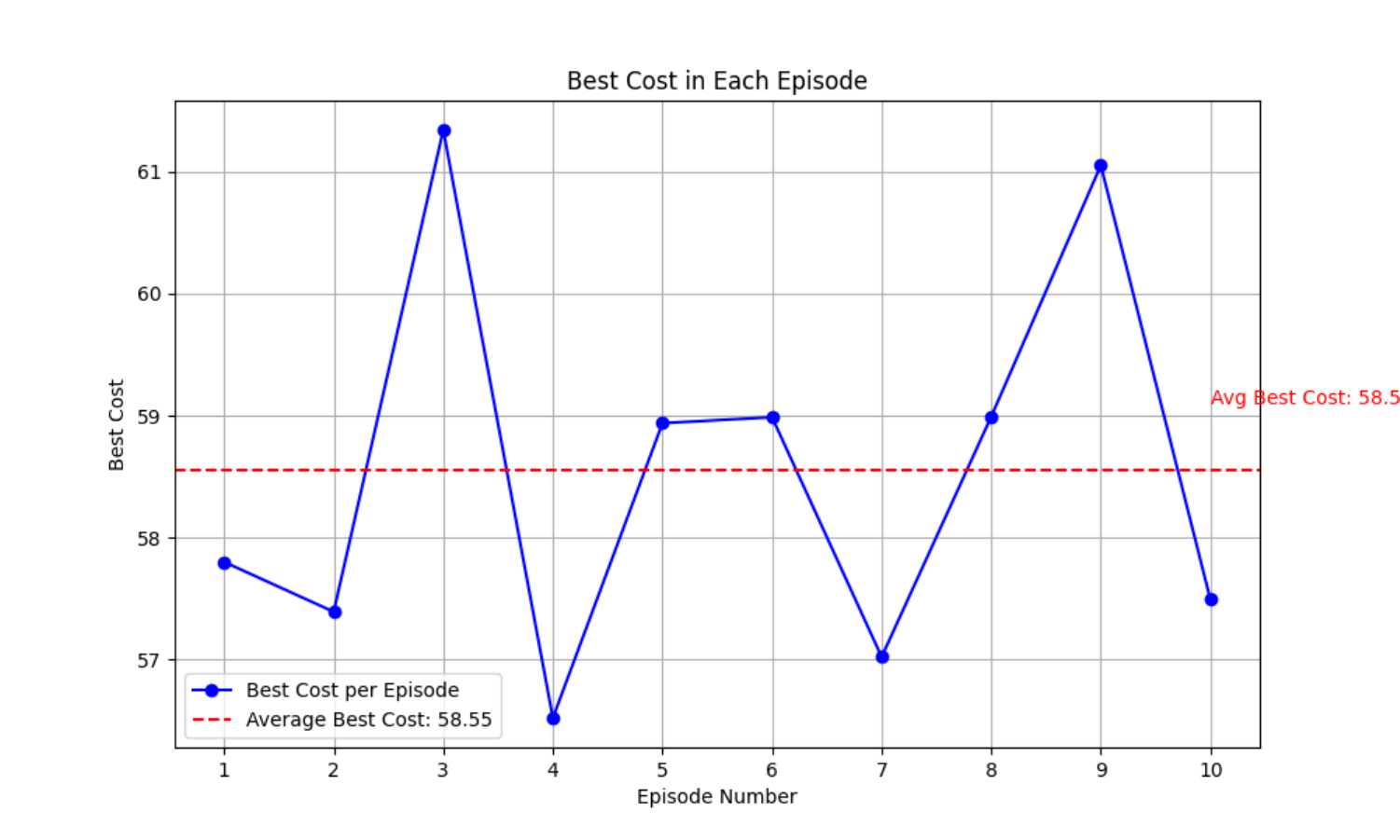}
    \caption{APSOSAC}
    \label{apso}
\end{figure}
\squeezeup
\subsection{Comparison and Analysis}

Our experimental results showed that the APSO-SAC method consistently achieved the lowest best costs compared to other methods. This indicates that integrating SAC with APSO allows dynamic adjustment of PSO parameters, enhancing both exploration and exploitation during the optimization process and leading to more efficient task offloading strategies.

\section{conclusion}

We proposed a novel method integrating Adaptive Particle Swarm Optimization (APSO) with Soft Actor Critic (SAC) reinforcement learning to enhance computational task offloading efficiency in Mobile Edge Computing (MEC) environments for IIoT applications. By dynamically adjusting PSO parameters, our approach effectively reduces latency and computational costs, enabling better exploration and convergence to optimal solutions. Experimental results demonstrate that the APSO-SAC method outperforms traditional PSO and other RL-integrated algorithms in handling large-scale, complex MEC environments, offering significant improvements in task-offloading efficiency crucial for IIoT systems.

Future work includes tuning the SAC hyperparameters to potentially achieve better results, which will require additional domain knowledge and extensive testing. Exploring discrete environmental models by focusing on smaller subsections may allow the use of other reinforcement learning algorithms such as Proximal Policy Optimization (PPO) or Advantage Actor-Critic (A2C). Further improvements to the simulator for finer tuning and specific advancements could also enhance the performance and applicability of our method.

\bibliographystyle{IEEEtran}

\end{document}